\documentclass[10pt,twocolumn,letterpaper]{article}
\pdfoutput=1
\usepackage{cvpr}
\usepackage{times}
\usepackage{epsfig}
\usepackage{graphicx}
\usepackage{amsmath}
\usepackage{amssymb}

\usepackage{algorithm}
\usepackage{caption}
\usepackage{subcaption}
\usepackage[noend]{algpseudocode}
\usepackage{enumitem}

\usepackage[breaklinks=true,bookmarks=false]{hyperref}

\cvprfinalcopy 

\def\cvprPaperID{****} 

\begin{document}

\title{\ Fashion-AttGAN: Attribute-Aware Fashion Editing with Multi-Objective GAN}

\author{Qing Ping, Bing Wu, Wanying Ding, Jiangbo Yuan\\
Vipshop US Inc.\\
San Jose, CA 95134, USA\\
{\tt\small \{pingqing508,bingwuthu,dingwanying0820,yuanjiangbo\}@gmail.com}
}

\maketitle

\begin{abstract}
  In this paper, we introduce attribute-aware fashion-editing, a novel task, to the fashion domain. We re-define the overall objectives in AttGAN~\cite{he2017attgan} and propose the Fashion-AttGAN model for this new task. A dataset is constructed for this task with 14,221 and 22 attributes, which has been made publically available. Experimental results show effectiveness of our Fashion-AttGAN on fashion editing over the original AttGAN.
\end{abstract}

\section{Introduction}

As we're witnessing great booming of online fashion shopping,
deep learning models such as GAN models\cite{goodfellow2014generative} have been widely applied to fashion industry, such as virtual try-on\cite{han2018viton}, domain adaptation\cite{almahairi2018augmented}, text2image \cite{rostamzadeh2018fashion} and so on. In this paper, we introduce a novel task into the fashion domain, namely attribute-aware fashion-editing, which edits certain attribute(s) of the image of a fashion item and preserve other details as intact as possible. This task opens a new door of possibilities for user-driven fashion design, and is potentially beneficial to virtual try-on, outfit recommendation, visual search and so on. Although previous work on human face editing is relevant to our proposed task \cite{perarnau2016invertible,he2017attgan,bao2017cvae,lu2018attribute},they are not directly applicable to fashion-editing, since now the scope of editing is not confined to a small area in human faces, but to much larger area of a fashion item, such as an entire sleeve or collar. To bridge these gaps, we present preliminary results of Fashion-AttGAN, a variant of face attribute-editing AttGAN. We define different overall objectives, and break the overly-strict constraint from the auto-encoder on the generator so that it can generate more ``wild" samples with dramatic changes.

The contributions of our work are three folds:
1) We introduce the task of fashion-editing into the fashion domain. To our best knowledge, this is the first time such task is established and explored in this area.
2) We adapt the original AttGAN to our task, and improve the model by re-defining the overall objectives.
3) We build a large-scale fashion dataset of 14,221 images and 22 attributes, which has been made publically available on \href{https://github.com/ChanningPing/Fashion_Attribute_Editing}{\textit{Github}}.

\section{Method}
Our Fashion-AttGAN model is a variant of previous  facial attribute-editing model AttGAN\cite{he2017attgan}. Both models include an encoder network $E$, a generator network $G$, a classification network $C$ and discriminator network $D$, where $C$ and $D$ share parameters except the last layer. For details of the structure of the model, please refer to AttGAN \cite{he2017attgan}. The main differences between our model and AttGAN are the overall objectives: we define three objectives of different optimization scopes:
\vspace{-1\baselineskip}

\begin{align}
\underset{\theta_{D},\theta_{C}}{min} \mathcal{L}_{D, C} =  \mathcal{L}_{{adv}_d} + \lambda_{2} \mathcal{L}_C(x_{\hat{a}})
\label{eq:1}\\
\underset{\theta_{E},\theta_{G}}{min} \mathcal{L}_{E, G} =  \mathcal{L}_{{adv}_g} + \lambda_{1} \mathcal{L}_{rec} 
\label{eq:2}\\
\underset{\theta_{G}}{min} \mathcal{L}_{G} =  \mathcal{L}_C(x_{\hat{b}})
\label{eq:3}
\end{align}

In AttGAN, the loss objectives in equation \ref{eq:2} and \ref{eq:3} are of the same scope for both encoder and decoder networks. The training details of Fashion-AttGAN is summarized in Algorithm \ref{alg:attFashion}.

The intuition behind our objective functions is that by back-propagating the classification error $\mathcal{L}_C(x_{\hat{b}})$ of attribute-edited generated sample $x_{\hat{b}}$ to as back as the generator network, the model is empowered with more flexibility to generate ``wild" samples related to different attribute edits, in contrast with propagating the errors back to encoders, which greatly limited the power of the generator network to generate novel samples, due to the auto-encoder path in AttGAN. This may not be a problem in face attribute-editing, since the editing is focused on small areas of human faces. However, when it comes to fashion-editing, original AttGAN does not allow generator sufficient flexibility to edit much larger areas, such as the length of the sleeves.

\captionsetup[subfigure]{skip=0pt} 
\begin{figure*}
\centering
\begin{subfigure}[b]{\textwidth}
   \includegraphics[width=\textwidth]{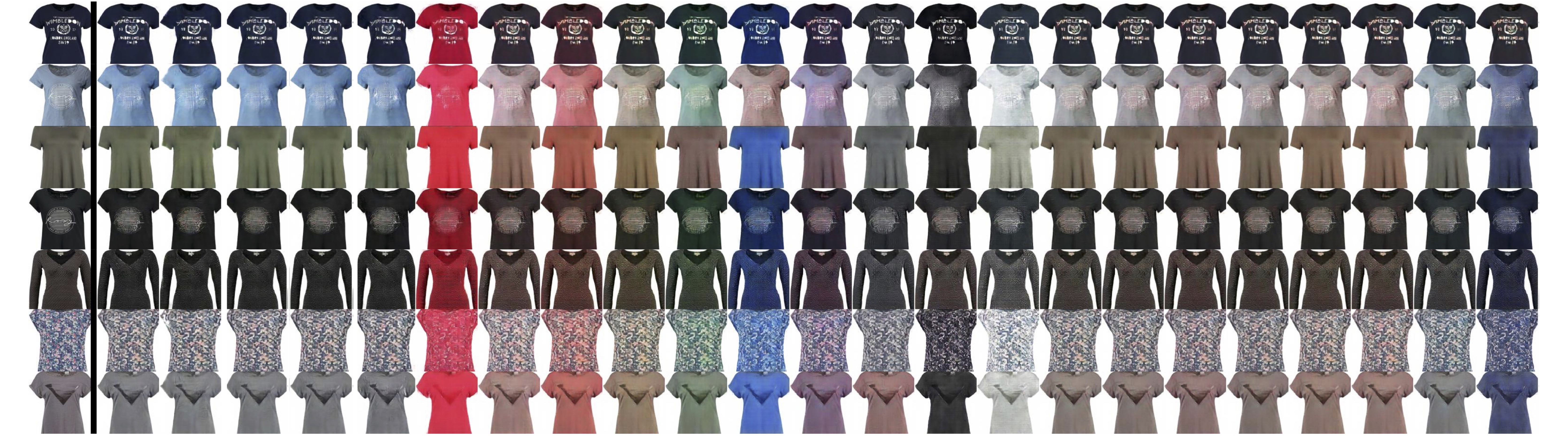}
   \caption{Attribute-Editing Results of AttGAN}
\end{subfigure}
\begin{subfigure}[b]{\textwidth}
   \includegraphics[width=\textwidth]{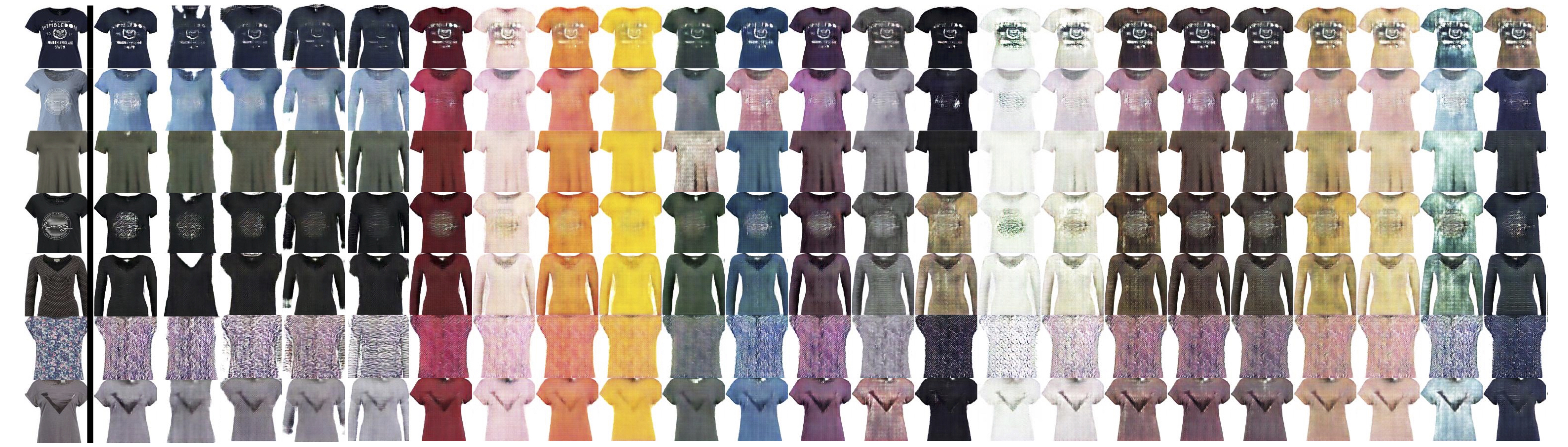}
   \caption{Attribute-Editing Results of Fashion-AttGAN}
\end{subfigure}
\vspace{-2\baselineskip}
\caption{Clothing Attribute-Editing Results. From left to right columns:(1) original image, (2) reconstructed image, (3-6) varied sleeve lengths, (7-24) varied colors.}
\label{fig:results}
\end{figure*}


\section{Experiments \& Results}


\subsection{Dataset}
In this study, we constructed a dataset based on the VITON dataset\cite{han2018viton,wang2018toward} which is publically available . Each entry in the dataset consists of an image from VITON, and a vector of attributes, such as ``no-sleeve", ``short-sleeve",``red",``blue" and so on. The attribute vector is predicted with our in-house classification model. The dataset includes 14,221 images, and 22 attribute values. In the future, we plan to publish a larger dataset with more images and more attributes.

\subsection{Experimental Results}
The comparative results of fashion-editing between our Fashion-AttGAN and AttGAN are depicted in Fig.\ref{fig:results}. From the figures we can observe that: (1) Color edits: the original AttGAN can edit colors of clothes with lighter shades, but not so well for darker shades (row:(1,4,5), column:(7-22)). Whereas our Fashion-AttGAN can modify clothes of almost any shade to other colors as shown in Fig.\ref{fig:results}(b); (2) Sleeve length: original AttGAN does not present any sleeve-length changes even with careful parameter tuning (Fig.\ref{fig:results}(a),column:(3,4,5,6)).On the contrary, our Fashion-AttGAN can generate vivid samples of different sleeve lengths and preserve the patterns in original samples (shapes,colors, logos, textures) as much as possible (Fig.\ref{fig:results}(b)).
\begin{algorithm}
\caption{The training pipeline of  Fashion-AttGAN }\label{alg:attFashion}
Require: ${\theta}_{E}$,${\theta}_G$,${\theta}_D$,${\theta}_C$ are initial image encoder network, generator network,  discriminator network  and attribute classification network parameters.  $\ell(\cdot)$ is a binary cross-entropy loss for an attribute.
\begin{algorithmic}[1]
\While{${\theta}$ has not converged}
\State {Sample ${x_a} \sim p_x$ a batch from real image data; }
\State {$z\leftarrow E(x_a)$ }
\State {$x_{\hat{a}}\leftarrow G(z, a)$ }
\State {$x_{\hat{b}}\leftarrow G(z, b)$ }
\State $\mathcal{L}_C(x_{\hat{a}}) \sim \mathbb{E}_{{x^a} \sim p_{data}}  [\sum_{i=1}^{n}\ell(a_i,C(x_{\hat{a}}))]$
\State $\mathcal{L}_C(x_{\hat{b}}) \sim \mathbb{E}_{{x^a} \sim p_{data}, b \sim p_{attr}}  [\sum_{i=1}^{n}\ell(b_i,C(x_{\hat{b}}))]$

\State $\mathcal{L}_{{adv}_d} \sim -\frac{1}{m}\sum_{i=1}^{m}[\mbox{log}D(x_a)+\mbox{log}(1-D(x_{\hat{b}})) $
\State $\mathcal{L}_{{adv}_g} \sim \frac{1}{m}\sum_{i=1}^{m}[\mbox{log}(1-D(x_{\hat{b}})) $

\State $\mathcal{L}_{rec} \sim \mathbb{E}_{{x^a} \sim p_{data}}  ||\Phi(x_a) - \Phi(x_{\hat{a}})||_{2}^{2}$

\State $ \theta_{D},\theta_C \stackrel{-}\leftarrow - \triangledown_{\theta_{D}, \theta_{C}} (\mathcal{L}_{{adv}_d} + \mathcal{L}_C(x_{\hat{a}}))$

\State $ \theta_{E},\theta_{G} \stackrel{-}\leftarrow - \triangledown_{\theta_{E},\theta_{G}} (\mathcal{L}_{{adv}_g}  + \mathcal{L}_{rec} )$

\State $ \theta_G \stackrel{-}\leftarrow - \triangledown_{\theta_{G}} (\mathcal{L}_C(x_{\hat{b}}))$
\EndWhile
\label{aclgandwhile}
\end{algorithmic}
\end{algorithm}


{\small
\bibliographystyle{ieee}

}

\end{document}